\documentclass[letterpaper, 10 pt, conference]{ieeeconf}  %

\IEEEoverridecommandlockouts                              %

\usepackage{graphicx} %
\usepackage{amsmath} %
\usepackage{amssymb}  %
\usepackage{amsbsy}

\usepackage{todonotes}
\usepackage{verbatim}

\usepackage{acronym}
\usepackage{overpic}
\usepackage[hyphens]{url}

\usepackage{algorithm}%
\usepackage{algpseudocode}%

\usepackage[utf8]{inputenc}

\usepackage{footnote}
\makesavenoteenv{tabular}
\makesavenoteenv{table}
\usepackage{cite}
\usepackage{subcaption}
\usepackage{booktabs}
\usepackage{tikz}
\usepackage{pgfplots}
\usepackage{xcolor}

\pgfplotsset{grid style={dotted, gray}}
\pgfplotsset{minor grid style={dotted,gray}}
\pgfplotsset{every tick label/.append style={font=\tiny}}
\pgfplotsset{every axis/.append style={font=\small}}
\pgfplotsset{ylabel near ticks}
\pgfplotsset{xlabel near ticks}
\newlength\figureheight 
\newlength\figurewidth

\newcommand{\figref}[1]{\hyperref[#1]{Figure~\ref*{#1}}}
\newcommand{\tabref}[1]{\hyperref[#1]{Table~\ref*{#1}}}
\newcommand{\secref}[1]{\hyperref[#1]{Section~\ref*{#1}}}
\newcommand{\algoref}[1]{\hyperref[#1]{Algorithm~\ref*{#1}}}

\newlength\myindent
\newcommand{\mbm}[1]{\mbox{\boldmath $#1$}}

\newcommand{\matr}[1]{\mathbf{#1}}

\def\ie{\textit{i.e.~}}
\def\eg{\textit{e.g.}}

\newacro{dmp}[DMP]{Dynamic Movement Primitives}
\newacro{gmm}[GMM]{Gaussian Mixture Model}
\newacro{gmr}[GMR]{Gaussian Mixture Regression}
\newacro{rl}[RL]{Reinforcement Learning}
\newacro{pomdp}[POMDP]{Partially Observable Markov Decision Process}
\newacro{tcp}[TCP]{Tool Center Point}
\newacro{tsp}[TSP]{Travelling Salesman Problem}
\newacro{lfd}[LfD]{Learning from Demonstration}
\newacro{pto}[PTO]{Projection-based Trajectory Optimization}
\newacro{em}[EM]{Expectation-Maximization}
\newacro{tshix}[TSHIX]{Trajectory Sampling from Human-Inspired Exploration}
\newacro{ftsensor}[F/T sensor]{force-torque sensor}
\newacro{vsa}[VSA]{Variable Stiffness Actuator}
\newacro{via}[VIA]{Variable Impedance Actuator}
\newacro{ml}[ML]{Maximum Likelihood}

\title{\LARGE \bf
Imitating Human Search Strategies for Assembly
}

\author{Dennis Ehlers\textsuperscript{1}, Markku Suomalainen\textsuperscript{1,2}, Jens Lundell\textsuperscript{1} and Ville Kyrki\textsuperscript{1}%
\thanks{This work was supported by Academy of Finland, decision 286580 and 
314180.}%
\thanks{\textsuperscript{1}D.\ Ehlers, J.\ Lundell and V.\ Kyrki are with School of Electrical Engineering, Aalto University, Finland {\tt\small firstname.surname@aalto.fi}}
\thanks{\textsuperscript{2}M.\ Suomalainen is currently with University of Oulu, Finland {\tt\small markku.suomalainen@oulu.fi}}
}

\begin{document}

\maketitle
\thispagestyle{empty}
\pagestyle{empty}

\begin{abstract}
We present a Learning from Demonstration method for teaching robots to perform search strategies imitated from humans in scenarios where alignment tasks fail due to position uncertainty. The method utilizes human demonstrations to learn both a state invariant dynamics model and an exploration distribution that captures the search area covered by the demonstrator. We present two alternative algorithms for computing a search trajectory from the exploration distribution, one based on sampling and another based on deterministic ergodic control. We augment the search trajectory with forces learnt through the dynamics model to enable searching both in force and position domains. An impedance controller with superposed forces is used for reproducing the learnt strategy. We experimentally evaluate the method on a KUKA LWR4+ performing a 2D peg-in-hole and a 3D electricity socket task. Results show that the proposed method can, with only few human demonstrations, learn to complete the search task. 
\end{abstract}

\section{INTRODUCTION}
\label{intro}
\begin{figure}
        \centering
        \begin{subfigure}[b]{0.225\textwidth}
            \centering
            \includegraphics[trim={0 3cm 0 1cm},clip,width=\textwidth]{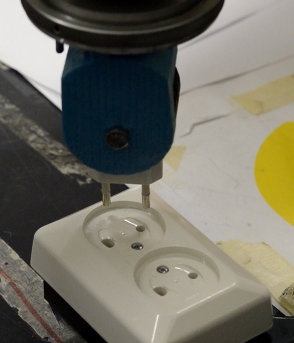}
        \end{subfigure}
        \hfill
        \begin{subfigure}[b]{0.225\textwidth}  
            \centering 
            \includegraphics[trim={0 3cm 0 1cm},clip,width=\textwidth]{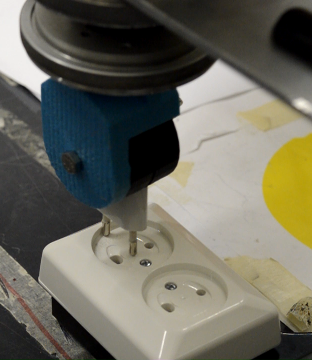}
        \end{subfigure}
        \vskip\baselineskip
        \begin{subfigure}[b]{0.225\textwidth}   
            \centering 
            \includegraphics[trim={0 3cm 0 1cm},clip,width=\textwidth]{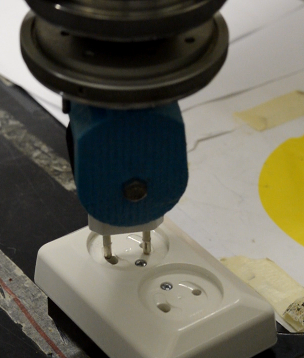}
        \end{subfigure}
        \quad
        \begin{subfigure}[b]{0.225\textwidth}   
            \centering 
            \includegraphics[trim={0 3cm 0 1cm},clip,width=\textwidth]{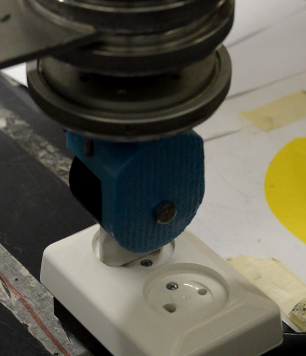}
        \end{subfigure}
        \caption{\small An example sequence of a robot inserting a plug into a socket without vision sensing or knowledge about the location of the socket's holes.} 
        \label{fig:plug_insertion}
        %\vspace{-1.5em}
    \end{figure}

To succeed in peg-in-hole type assembly tasks with small clearance between workpieces, a robot's internal model of the world must match with the environment to high precision, as even small errors in the model can cause the alignment to fail.
It is possible to rely on handcrafted exception strategies to 
complete the task, but the process of manually defining the 
strategies is tedious and often not generalizeable.
Even though recent research has shown significant improvement in planning of contact motions \cite{guan2018planning}, they are ineffective when facing modeling errors.
Our key insight is to make the robot behave similar to a human when faced with such uncertainties about the environment---probing the environment, like fitting a key into a lock in darkness or inserting a plug into an electric socket, as illustrated in Fig.~\ref{fig:plug_insertion}.
Thus we present a \ac{lfd} method for teaching robots to perform human-inspired search strategies.

In \ac{lfd}, a human teacher performs one or more demonstrations that the robot 
then learns to mimic \cite{argall2009survey}. Methods to encode and 
parametrize the demonstrated skill are usually based on attractors along a 
trajectory, such as \ac{dmp} \cite{schaal2006dynamic} and \ac{gmm} with 
\ac{gmr} \cite{calinon2007learning}. However, these methods assume that the 
goal pose is either known or can be learned. In contrast, we target scenarios 
where the goal pose can vary between reproductions of the same task and cannot 
be visually identified with sufficient accuracy to complete the task, thus rendering the problem closer to area coverage than a trajectory following problem.

We propose a method to learn search strategies by imitating human behavior 
in scenarios where the transformation between the start and 
end poses is not known but the human follows a deliberative strategy to cover an area that contains the end pose. The method 
consists of 
four steps: 1) gathering one or more human demonstrations from which forces and 
positions are recorded; 2) modeling the covered search area---the 
\textit{exploration distribution}---and state-invariant dynamics of the task at 
hand probabilistically; 3) planning a search trajectory over the exploration distribution; and 4) 
executing the search trajectory with an impedance controller while 
superposing forces learned from dynamics. For planning the search trajectory 
we compare two algorithms: an ergodic control approach proposed by Miller et 
al.~\cite{miller2013trajectory} and a new sampling-based algorithm 
\ac{tshix}. 
We evaluate our method in both 2D (round peg-in-hole) 
and 3D (electric plug insertion) search tasks.

The main contributions in this work include 1) a novel 4-step framework  for learning in-contact search strategies from human demonstrations, 2) a sampling based method, \ac{tshix}, for creating trajectories from an exploration distribution and 3) showing through experiments that our learned search strategies are suitable to work as exception strategies in a 3D assembly task.

\section{RELATED WORK}
\label{RELATED}

Existing work on search motions mainly focus on exception strategies on a 2D plane. Abu-Dakka et al.~\cite{abu2014solving} presented an exception strategy based on stochastic search for failure case in a peg-in-hole task. However, their method did not use any task-specific information, which could improve the accuracy. Jasim et al.~\cite{jasim2014position} used an Archimedean spiral to cover the 2D search area. However, the approach is not applicable in higher dimensional search tasks such as inserting a plug into a socket. Baum et al.~\cite{baum2017opening} compared different physical exploration strategies in a 3D scenario. However, they did not learn the search strategies from human demonstrations. 

A similar idea of using human demonstrations to learn search strategies was previously explored by De Chambrier in \cite{de2016learning}. To learn to insert a plug into a socket, the method used 150 demonstrations to train a \ac{pomdp} which learned a strategy that first localized itself in the environment with the help of physical barriers, and then followed a fixed trajectory to the socket. In comparison, our method focuses on cases with known approximate location of the goal, while the position or orientation of the goal can change from one demo to another in an unknown manner, requiring physical exploration for insertion. Moreover, our method does not require such an extensive amount of demonstrations.

In De Chambrier's work, the actual insertion was performed according to the work of Kronander \cite[Chapter 5]{kronander2015control}.
He also had an underlying hand-crafted strategy based on contact transients, and in case of jamming the human teacher intervened.
In contrast, our method can learn the whole search process from humans in cases where the approximate location and orientation of the hole is known.
In addition, our search strategy considers both position and force distributions, thus allowing also force-based search.
Furthermore, the presented method does not require hand-crafting transient behaviors.

\section{METHOD}
\label{sec:METHOD}

Our proposed method for learning search strategies from human demonstrations 
consist of 4 steps: 1) learning the exploration distribution 
Section~\ref{sec:exploDist}, 2) creating the search trajectory for the 
manipulator Section~\ref{sec:trajectory}, 3) learning the task dynamics 
Section~\ref{sec:dynamics}, and 4) performing the search trajectory with an 
impedance controller Section~\ref{sec:searchPerform}.

\subsection{Learning the Exploration Distribution}
\label{sec:exploDist}

We model the exploration distribution with a Gaussian\footnote{For more 
complex exploration patterns, the single Gaussian can be replaced with a 
\ac{gmm}.} distribution over the task space,
%, using \ac{ml} Estimation:
\begin{equation}\label{eq:exploDist_Gauss}
    p_{e}(\mbm{s}) = \mathcal{N} (\mbm{s}|\mbm{\mu},\matr{\Sigma})\ ,
\end{equation}
where $\mbm{s} \in \mathbb{R}^D$, $D$ is the dimensionality of the task (e.g. 3 if only translations are considered), and $\mbm{\mu}$ and $\matr{\Sigma}$ are the mean and covariance estimated by maximum likelihood. 

To learn expressive exploration distributions, the training data must represent
the entire search domain.
In this work the humans demonstrated the movement with closed eyes to mimic the sensing modalities of the robot.
To start a demonstration the end-effector was first placed in the proximity of the goal that resembles possible failure scenarios and then the human performed the demonstrations with his eyes closed.
Demonstrations that ended either immediately at the goal or in a configuration from which success without vision was improbable were discarded.

The demonstrations were aligned with respect to their starting positions, termed the \textit{search frame}.
With this choice we can learn the \textit{search} strategy of the teacher in situations where localizing the tool w.r.t.~the world frame is impossible. 
Thus any trajectory generated will also start at the origin of the search frame, making the learned search strategies independent of their location in the world frame.

\subsection{Creating the Search Trajectory}
\label{sec:trajectory}
We propose two methods for generating a search trajectory that covers the modeled exploration distribution: 1) \ac{tshix}, a stochastic sampling-based method, and 2) ergodic control \cite{miller2013trajectory}, a deterministic method. 

\subsubsection{Trajectory Sampling from Human-Inspired eXploration}
\label{sec:tsp}
\ac{tshix} samples points from the exploration distribution created in~\ref{sec:exploDist}, 
finds an approximation of the shortest itinerary through the sampled points and smooths it. 

We draw samples from the exploration distribution using Marsaglia's ziggurat method~\cite{marsaglia1984fast}.
Visualizations of sampling in 2D are presented in Fig.~\ref{fig:PiH_samp}.
Finding the shortest itinerary through the sampled points is a \ac{tsp}, which we solved using an existing genetic algorithm based optimizer for simplicity of implementation\footnote{\url{https://se.mathworks.com/matlabcentral/fileexchange/21198-fixed-start-open-travelling-salesman-problem-genetic-algorithm}}.
The start of the itinerary is always set at the origin of the search frame in an effort to imitate the human search.
The itinerary is then smoothed with a Savitzky-Golay Filter \cite{savGolFilter}.
Fig.~\ref{fig:PiH_TSP_comp} illustrates the stages of creating a trajectory for the tool: A series of sample points are generated, through which an itinerary is calculated and smoothed to produce the final trajectory.

\subsubsection{Ergodic Control}
\label{sec:ergodic}
Ergodic control is a method for creating a trajectory such that at every step 
the time-averaged behavior of the system matches a known desired 
probability distribution. This is realized by minimizing a cost function that matches spatial statistics of the trajectory to those of the desired distribution. We quantify this as matching their Fourier coefficients \cite{mathew2011metrics}
\begin{equation}
\label{eq:ergodic}
  C\big(x(t)\big) = \sum_{k=1}^K \Lambda_{k} \Big( c_k\big(x(t)\big) - \phi_k \Big)^{\!2},
\end{equation}
where $\Lambda_k$ are frequency dependent weights, and $\phi_k$ are the Fourier coefficients of the exploration distribution learnt from the human demonstration.
To minimize (\ref{eq:ergodic}), we use the optimal control formulation of
Miller and Murphey~\cite{miller2013trajectory}. 

\subsection{Learning the Task Dynamics}
\label{sec:dynamics}
When humans perform search tasks they do not simply follow a trajectory in space, but maintain contact between the workpieces and deduce vicinity of alignment based on contact forces.
Learning the dynamics is useful to efficiently overcome friction as well as to reach threshold forces for phenomena such as "snapping" \cite{stolt2011force}. Therefore, learning search strategies similar to humans requires simultaneous learning of positions and forces.

For this reason, to record contact forces during a demonstration a \ac{ftsensor} was attached to the wrist of the manipulator to record the interaction force (normal plus friction force).
Because the pose of the goal is not known, we learn location-invariant dynamics that we assume to be applicable for the entire search.

We model the dynamics as a joint Gaussian distribution of velocities and wrenches
\begin{equation}\label{eq:multivarGauss_locInv}
    p_{d}(\Delta \mbm{s}, \mbm{a}) = \mathcal{N} (\Delta \mbm{s}, \mbm{a} | \mbm{\mu},\matr{\Sigma})\ ,
\end{equation}
where $\Delta \mbm{s}_{t} = \mbm{s}_{t+1}- \mbm{s}_t$ is the instantaneous motion and $\mbm{a}_{t} = [ -\mbm{f}_{t}^T\ -\mbm{t}_{t}^T ]^T$ is the action wrench (inverse of the opposing measured interaction force).
The parameters $\mbm{\mu}$ and $\matr{\Sigma}$ are fitted to demonstrations using maximum likelihood.

We can then predict the required action wrench for a desired state change by conditioning the Gaussian distribution \eqref{eq:multivarGauss_locInv} on the state change.
Partitioning $\mbm{\mu} = [ \mbm{\mu_{a}}^T\,\ \mbm{\mu_{s}}^T ]^T$ and 
\begin{equation}\label{eq:conditionalGauss3}
\matr{\Sigma} = \begin{bmatrix}
\matr{\Sigma_{aa}} & \matr{\Sigma_{as}}\\
\matr{\Sigma_{sa}} & \matr{\Sigma_{ss}}
\end{bmatrix}\ ,
\end{equation}
the predicted action wrench for a desired state change $\Delta \mbm{s}^*$ is
\begin{equation}\label{eq:conditionalGauss4}
    ( \mbm{a}^*\ |\ \Delta \mbm{s}^* ) \sim \mathcal{N} (\mbm{\mu^*}, \matr{\Sigma^*})\ ,
\end{equation}
where
\begin{equation}\label{eq:conditionalGauss5}
    \mbm{\mu}^* = \mbm{\mu_{a}} + \matr{\Sigma_{as}} \matr{\Sigma_{ss}^{-1}}(\Delta \mbm{s}^* - \mbm{\mu_{s}})\ ,
\end{equation}
and
\begin{equation}\label{eq:conditionalGauss6}
    \matr{\Sigma}^* = \matr{\Sigma_{aa}} - \matr{\Sigma_{as}} \matr{\Sigma_{ss}^{-1}} \matr{\Sigma_{sa}}\ .
\end{equation}

\subsection{Performing the Search}
\label{sec:searchPerform}
To reproduce the search motions, we use an impedance controller with superposed forces and torques, illustrated in Fig.~[ref{fig:impedanceControl}.
We use an impedance controller to enable the robot to follow a trajectory in free space as well as in contact while being compliant to the environment.
Using the superposed feed-forward forces enables force based searching, provided the teacher demonstrates such behavior.

To determine the position set-points, the continuous search trajectory (Sec.\ III-C) is first sampled in time.
For each sample, the desired action wrench $\mbm{a}^*\equiv [\mbm{f}_f^T \ \mbm{t}_f^T]^T$ is calculated as the mean $\mbm{\mu}^*$ of the conditional distribution given in \eqref{eq:conditionalGauss5}.
The resulting impedance control law for translations is then
\begin{equation}\label{eq:impedanceControlForce}
   \mbm{f}_{c} =  K_{p} \mbm{e}_{p} + B_{p} \Dot{\mbm{e}}_{p} + \mbm{f}_{f}\ ,
\end{equation}
where $\mbm{f}_c$ is the Cartesian force commanded to the robot, $\mbm{e}_{p}$ is the positional error between current and desired position, $K_{p}$ the stiffness matrix, $\Dot{\mbm{e}}_{p}$ the derivative of the positional error, $B_{p}$ is the positional damping coefficient and $\mbm{f}_{f}$ the superposed feed-forward force. 

For the orientation, the impedance control law can be written
\begin{equation}\label{eq:impedanceControlTorque}
    \mbm{t}_{c} =  K_{\theta} \mbm{e}_{\theta} + B_{\theta} \Dot{\mbm{e}}_{\theta} + \mbm{t}_{f}\ ,
\end{equation}
where $\mbm{t}_c$ is the Cartesian torque command, $\mbm{e}_{\theta}$ is the angular error between current and desired orientation, $K_{\theta}$ is the angular stiffness matrix, $\Dot{\mbm{e}}_{\theta}$ is the derivative of the angular error, $B_{\theta}$ is the angular damping matrix and $\mbm{t}_{f}$ the superposed feed-forward torques.

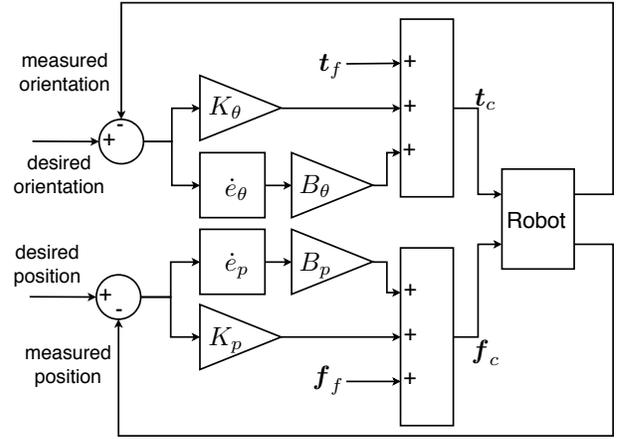
\begin{figure}
\begin{overpic}[width=1.0\columnwidth,tics=10]{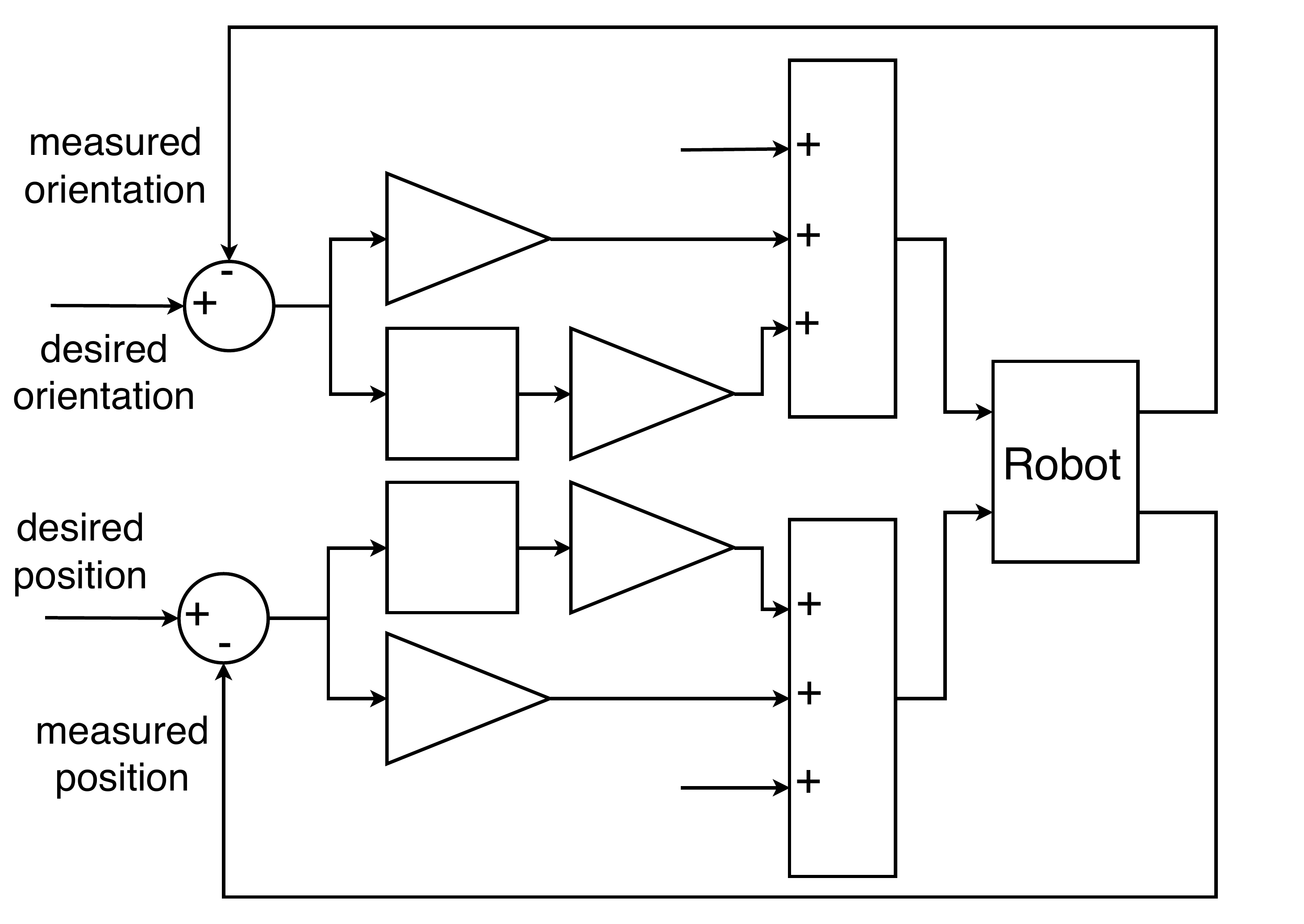}
    \put(31, 51.5){$K_{\theta}$}
    \put(33.5, 39){$\dot{e}_{\theta}$}
    \put(45, 39.5){$B_{\theta}$}
    \put(72, 53){$\mbm{t}_{c}$}
    \put(48, 58.25){$\mbm{t}_{f}$}
        
    \put(31, 16){$K_{p}$}
    \put(33.5, 27.5){$\dot{e}_{p}$}
    \put(45, 27.5){$B_{p}$}
    \put(71.5, 14){$\mbm{f}_{c}$}
    \put(47, 10){$\mbm{f}_{f}$}
\end{overpic}
\caption{\small{An impedance controller with F/T feed-forward is used to reproduce the search motions.}}
\label{fig:impedanceControl}
%\vspace{-1em}
\end{figure}

\section{EXPERIMENTS AND RESULTS}
\label{sec:experiments}
We used a KUKA LWR4+ lightweight arm to experimentally validate our method. Demonstrations were recorded in gravity compensation mode, where the robot's internal sensors recorded the pose of the robot and an ATI mini45 \ac{ftsensor} at the wrist of the robot recorded the wrench. We implemented the controller through the Fast Research Interface (FRI)~\cite{schreiber10} similarly as in \cite{suomalainen2016,suomalainen2017} such that KUKA's internal controller handles the feed-forward dynamics on top of the impedance controller defined in (\ref{eq:impedanceControlForce}-\ref{eq:impedanceControlTorque}).

\begin{figure}
    \centering
    \begin{subfigure}[t]{0.45\columnwidth}
        \includegraphics[width = \columnwidth]{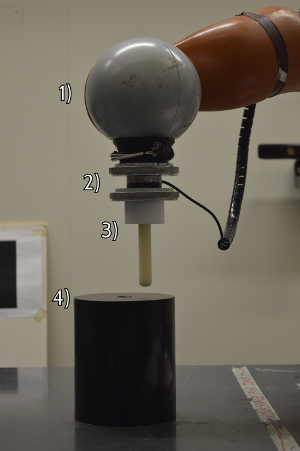}
        \caption{\small{Peg-in-Hole setup: 1) KUKA LWR4+, 2) \ac{ftsensor}, 3) Peg tool, 4) Cylinder with hole}}
        \label{fig:expSetupPih}
    \end{subfigure}
    ~           \begin{subfigure}[t]{0.45\columnwidth}
        \includegraphics[width = \columnwidth]{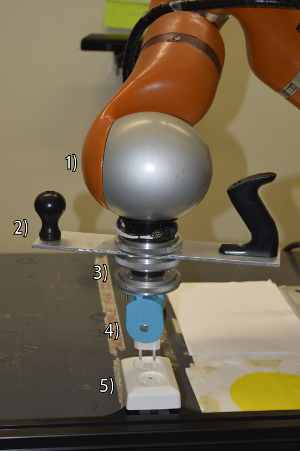}
        \caption{\small{The plug and socket setup: 1) KUKA LWR4+, 2) Grab-handle for users to manipulate the robot, 3) \ac{ftsensor}, 4) Plug tool, 5) Socket}}
        \label{fig:expSetupPlug}
    \end{subfigure}
    \caption{\small{The setups used for the experiments.}}\label{fig:expSetup}
    %\vspace{-1.5em}
\end{figure}

We evaluated the method in two scenarios: a peg-in-hole task (Fig.~\ref{fig:expSetupPih}) and an electrical plug and socket task (Fig.~\ref{fig:expSetupPlug}). With the peg-in-hole setup, we first compared the two presented trajectory creation algorithms, \ac{tshix} and ergodic control (Section~\ref{sec:tspVsErgodic}), then examined how the number of itinerary points for \ac{tshix} affected success rate (Section~\ref{sec:samplingSize}), and finally analyzed the trajectory following performance of the controller (Section~\ref{sec:feedForward}) and whether involving the dynamics in a 2D task improves the performance (Section~\ref{sec:dynamics}). With the electrical plug task (Section~\ref{sec:3Dsearch}) we study the performance of the method in a more complex 3D search task.

\subsection{Comparison Between Ergodic Control and \ac{tshix}}
\label{sec:tspVsErgodic}
We used the 2D peg-in-hole task (Fig.~\ref{fig:expSetupPih}) to compare the sampling based \ac{tshix} and ergodic control approaches for trajectory creation\footnote{To generate ergodic control trajectories we used the 
following open source implementation 
\url{https://github.com/dressel/ErgodicControl.jl}.}. We first learned the 
exploratory distribution and
the dynamics from a single demonstration. 
Since the peg-in-hole task is conducted in the 2D plane on top of 
a cylinder, the orientation of the peg tool was fixed during both demonstration 
and reproduction. With \ac{tshix} we generated 30 different trajectories in the 
x-y-plane, each consisting of 300 points sampled from the learned exploration 
distribution. With the ergodic control we generated a trajectory that was approximately the same length as the mean of the \ac{tshix} trajectories. Finally, for the trajectories 
provided by both algorithms we used the learned dynamics model to compute the 
superposed forces. 
 
We performed the comparison by randomly choosing 30 starting positions from within a circle of ca.~3~cm in diameter, located roughly halfway between the originally demonstrated starting position  and the hole of the cylinder. The reason for moving the starting positions closer to the hole is that the demonstration acts as a worst case scenario, \ie the one end of the exploration distribution will be approximately at the starting position. The same exploratory distribution and dynamics were used for both algorithms. For \ac{tshix} we used a different randomly chosen trajectory with each starting position to average the inherent randomness, whereas for the ergodic control we used the same trajectory with each position, due to its deterministic nature. As this task does not involve rotations, the control was performed by feeding the trajectory and superposed forces to~\eqref{eq:impedanceControlForce}.

The results of the experiment are presented in Table~\ref{tab:resultTspVsErgodic}. They show that \ac{tshix} performed 
slightly better than the ergodic control. In the five trials where \ac{tshix} failed, the ergodic control failed as well. 
These failures seemed to occur because the starting positions were too far away from the goal, as 
neither trajectory lead close enough to the hole. In the five trials where \ac{tshix} succeeded and ergodic control failed, 
the ergodic trajectory would usually go within the vicinity of the hole, but not cover it. This implies that the difference 
between the algorithms could stem from density of coverage. 

\begin{table}[t]
    \centering
    \begin{tabular}{l c c}
        \toprule
        Algorithm & Ergodic & \ac{tshix}\\
        \midrule
        Success rate & 20/30 & 25/30 \\
        Success percentage & 67\% & 83\% \\
        \bottomrule
    \end{tabular}
    \caption{\small{Success rates from peg-in-hole experiments from the two trajectory creation methods.}}
    \label{tab:resultTspVsErgodic}
    %\vspace{-1.2em}
\end{table}

One advantage of using the sampling-based \ac{tshix} is the flexibility in generating the trajectory. If a generated 
trajectory was unsuccessful or undesirable for other reasons, a new trajectory can easily be generated by changing the 
random seed for the sampling. Moreover, additional points could even be sampled during reproduction should the task fail. This is 
not possible for the ergodic trajectory, as they are per definition deterministic and thus when the chosen convergence 
criteria are met, the trajectory is finished. However, the ergodic trajectory can be considered optimal for covering a 
distribution~\cite{dressel2018optimality} and the deterministic nature increases the predictability and repeatability of 
the algorithm.

\subsection{Effect of Sample Size}
\label{sec:samplingSize}
\begin{figure}
    \centering
    \begin{subfigure}[tbp]{0.4\columnwidth}
        \includegraphics[width = \columnwidth]{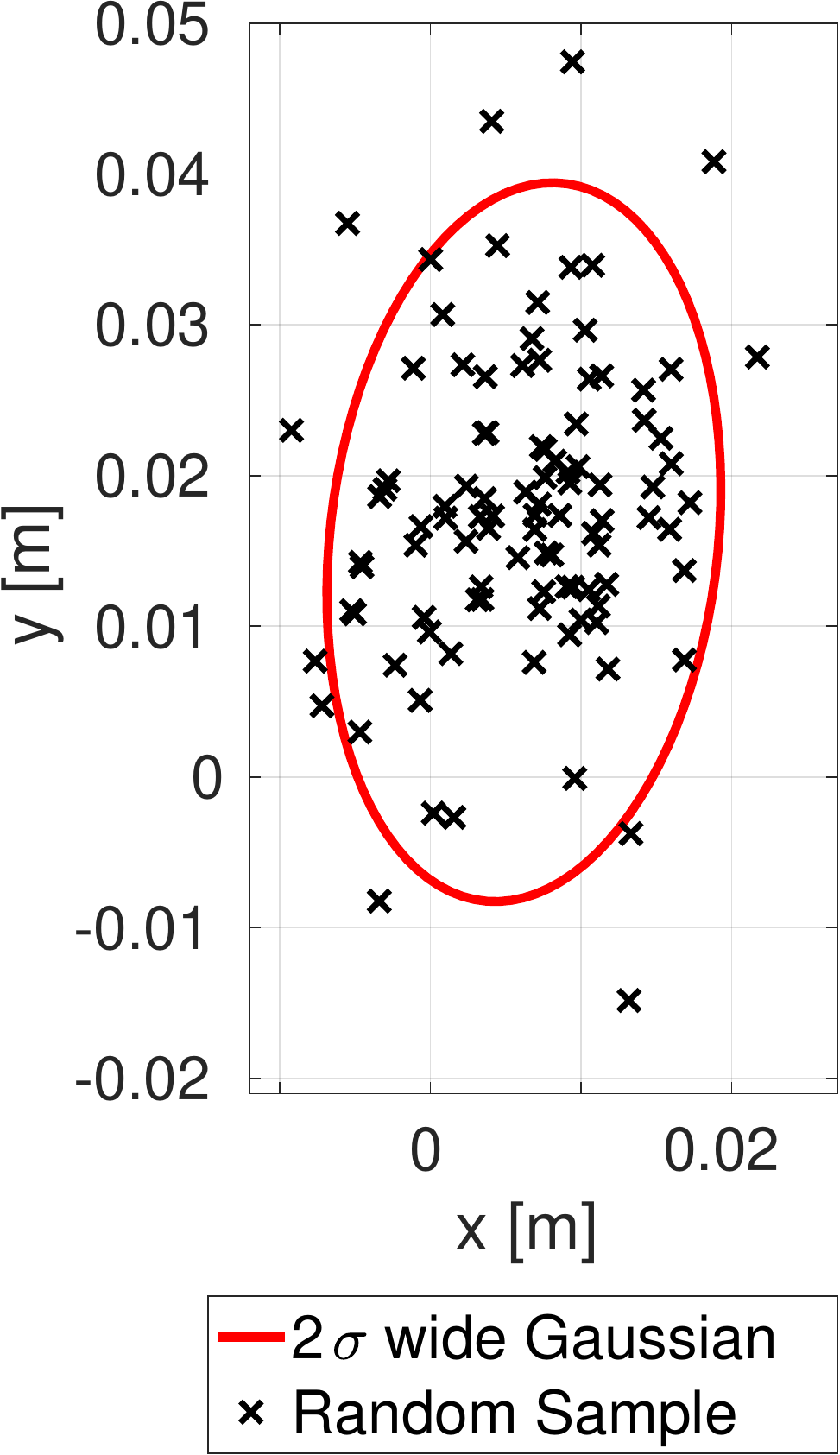}
   		\caption{\small{100 samples}}
    		\label{fig:PiH_samp100}
    \end{subfigure}
    ~           \begin{subfigure}[tbp]{0.4\columnwidth}
        \includegraphics[width = \columnwidth]{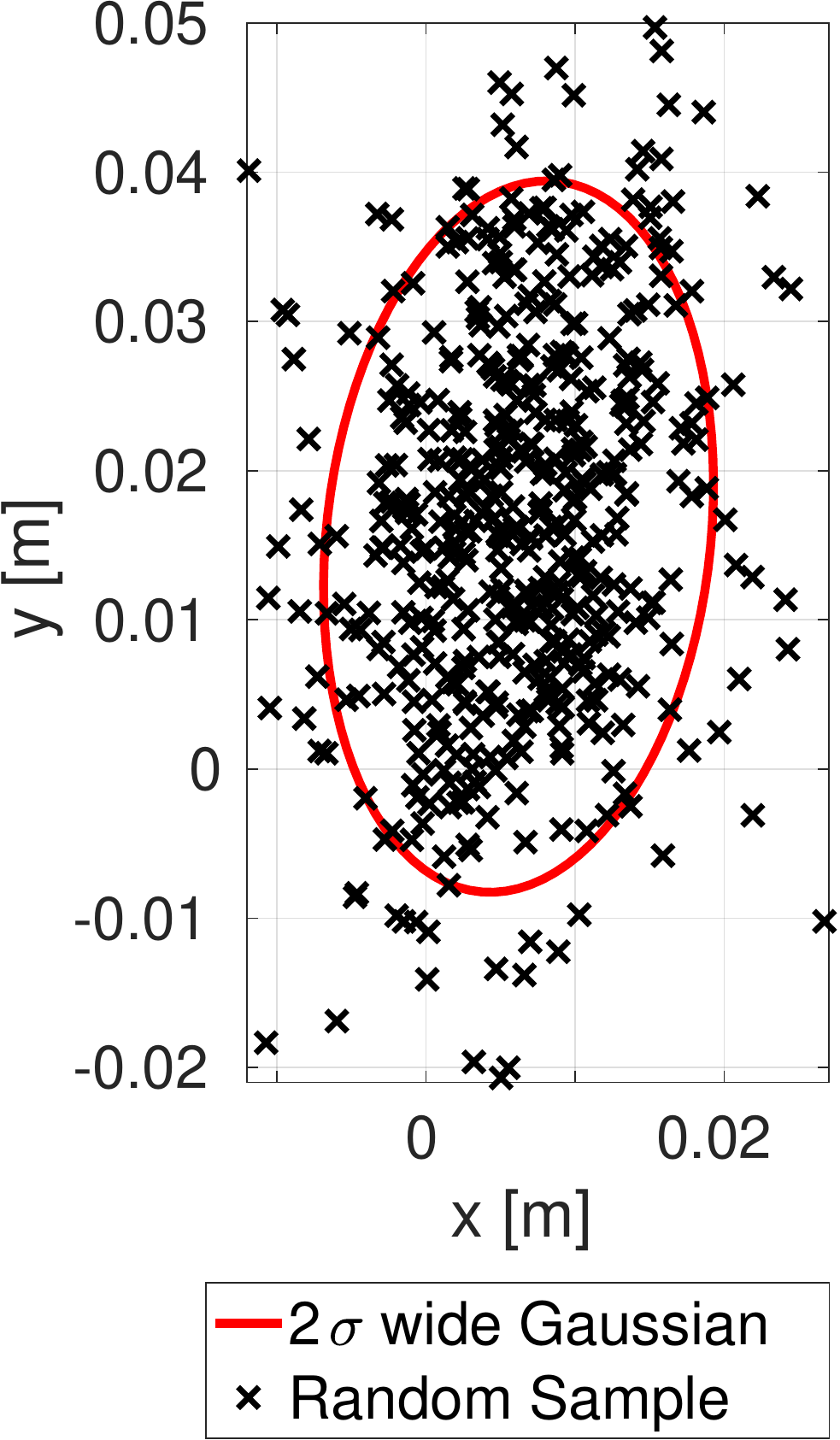}
   		\caption{\small{500 samples}}
    		\label{fig:PiH_samp500}
    \end{subfigure}
   	\caption{\small{Different number of sampled points covering the learned exploration distribution.}} \label{fig:PiH_samp}
   	%\vspace{-1.5em}
\end{figure}

To investigate the effect the sample size had on the success rate of \ac{tshix}, one successful starting position from the 
previous section was selected for further analysis. We generated 20 trajectories for sample sizes of 100, 200, 300, 400 and 
500 each. More samples will lead to a higher coverage (see Fig.~\ref{fig:PiH_samp}) and the connection between coverage and 
success rate is investigated in this section. The trajectories and their corresponding predicted superposed forces were 
executed with the impedance controller from~\eqref{eq:impedanceControlForce}. The results, presented in 
Table~\ref{tab:samplingSizeComp}, show that initially the success of the search 
increases with higher sample sizes but so does the 
computational time. 
However, somewhat counterintuitively, after 300 samples the success rate consistently stays at 85\%, a finding further explored in the next section.

\begin{table}[h]
	\begin{tabular}{l c c c c c}
		\toprule
		Sample size                                                               & 100  & 200   & 300   & 400   & 500   \\
		\midrule
		Success rate                                                              & 6/20 & 12/20 & 17/20 & 17/20 & 17/20 \\
		Success percentage                                                        & 30\% & 60\%  & 85\%  & 85\%  & 85\%  \\
\begin{tabular}[c]{@{}l@{}}Mean CPU time (s)\footnote{In this work all computations were performed using 
		an Intel(R) Core(TM)2 Quad CPU Q9400 @ 2.66GHz.}\\ ($\pm 1$ sd) 
(s)\end{tabular} & 
\begin{tabular}[c]{@{}l@{}}13.8\\ ($\pm$0.3)\end{tabular} & 
\begin{tabular}[c]{@{}l@{}}34.1\\ ($\pm$1.1)\end{tabular} & 
\begin{tabular}[c]{@{}l@{}}61.9\\ ($\pm$4.4)\end{tabular} & 
\begin{tabular}[c]{@{}l@{}}94.3\\ ($\pm$5.4)\end{tabular} & 
\begin{tabular}[c]{@{}l@{}}135.6\\ ($\pm$10.4)\end{tabular} \\
        \bottomrule
	\end{tabular}
    \caption{\small{Success rates and computational times from varying the number of sampled points in \ac{tshix} on the 2D peg-in-hole task }}
    \label{tab:samplingSizeComp}
    %\vspace{-0.5em}
\end{table}

\subsection{Controller Analysis}
\label{sec:feedForward}
For in-contact tasks it is a challenge to choose the right stiffness value such that the trajectory is followed while allowing compliance to avoid excessive contact forces and facilitate easier completion of the task. For peg-in-hole tasks, lower stiffness perpendicular to the motion allows the peg to be inserted without moving over the exact center of the hole as it exploits the mechanical gradients of the task. 

We chose to keep the stiffness low and instead provide the robot with the dynamics of the task. By knowing the dynamics instead of increasing the stiffness, we can predict the forces necessary to counter the environmental forces, such as friction, along the trajectory. The stiffness is therefore increased only in the direction of motion while leaving the tool compliant perpendicular to the direction of motion. We empirically observed that the perpendicular compliance indeed seemed to play a key role in the final stage of the task where the peg has to slide into the hole; without the compliance the peg was more likely to slide by close to the hole but not sink in.

To experimentally examine how much influence the usage of the superposed forces has on the outcome of the task, an additional experiment series was set up. 20 trajectories from 100, 200 and 400 sampled points were created. Each trajectory was then executed with the impedance controller in Section~\ref{sec:samplingSize} without the addition of the superposed predicted forces. For the 100 sample point trajectories, only 2/20 trials succeeded, 4 less then with force-feed-forward. Similar observations were made for 200 and 400 sample point trajectories (7/20 and 13/20 respectively). This consistently inferior performance is indicative of the importance of the superposed forces even on a task performed on a 2D plane. As the forces are expected to be countering the environmental forces, which in this task consist of mainly friction, the forces point in the direction of the state change. 
This is observable in Fig.~\ref{fig:PiH_predictedFrc} where the superposed forces are plotted over a trajectory sliding 
along a 2D plane. We note that in this example the forces pointing "north-west" have smaller magnitude and are more 
difficult to observe. 

\begin{figure}
    \centering
    \begin{subfigure}[tbp]{0.48\columnwidth}
	\includegraphics[width = \columnwidth]{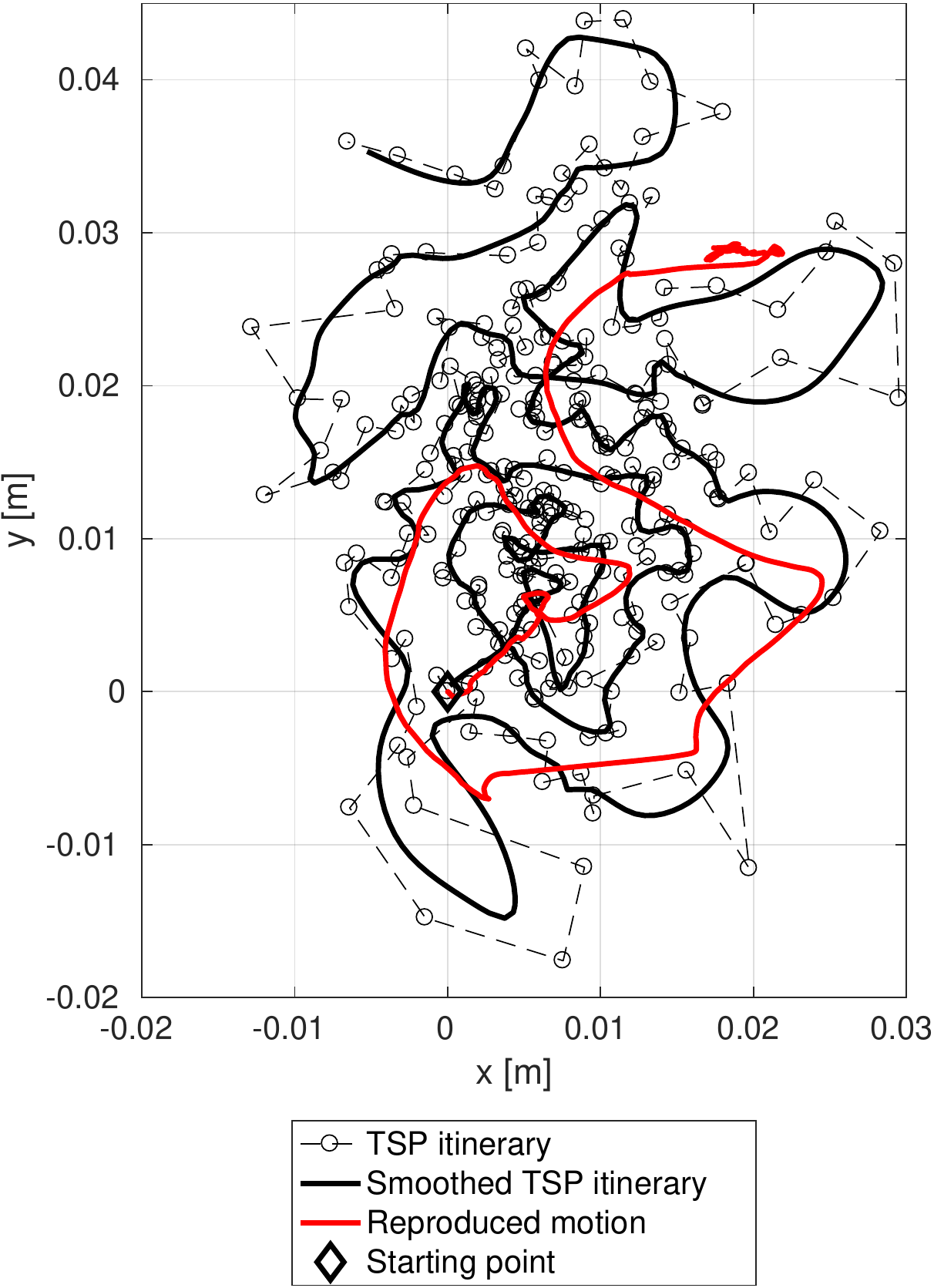}
   		\caption{}    		\label{fig:PiH_TSP_comp}
    \end{subfigure}
    ~           \begin{subfigure}[tbp]{0.48\columnwidth}
        \includegraphics[width = \columnwidth]{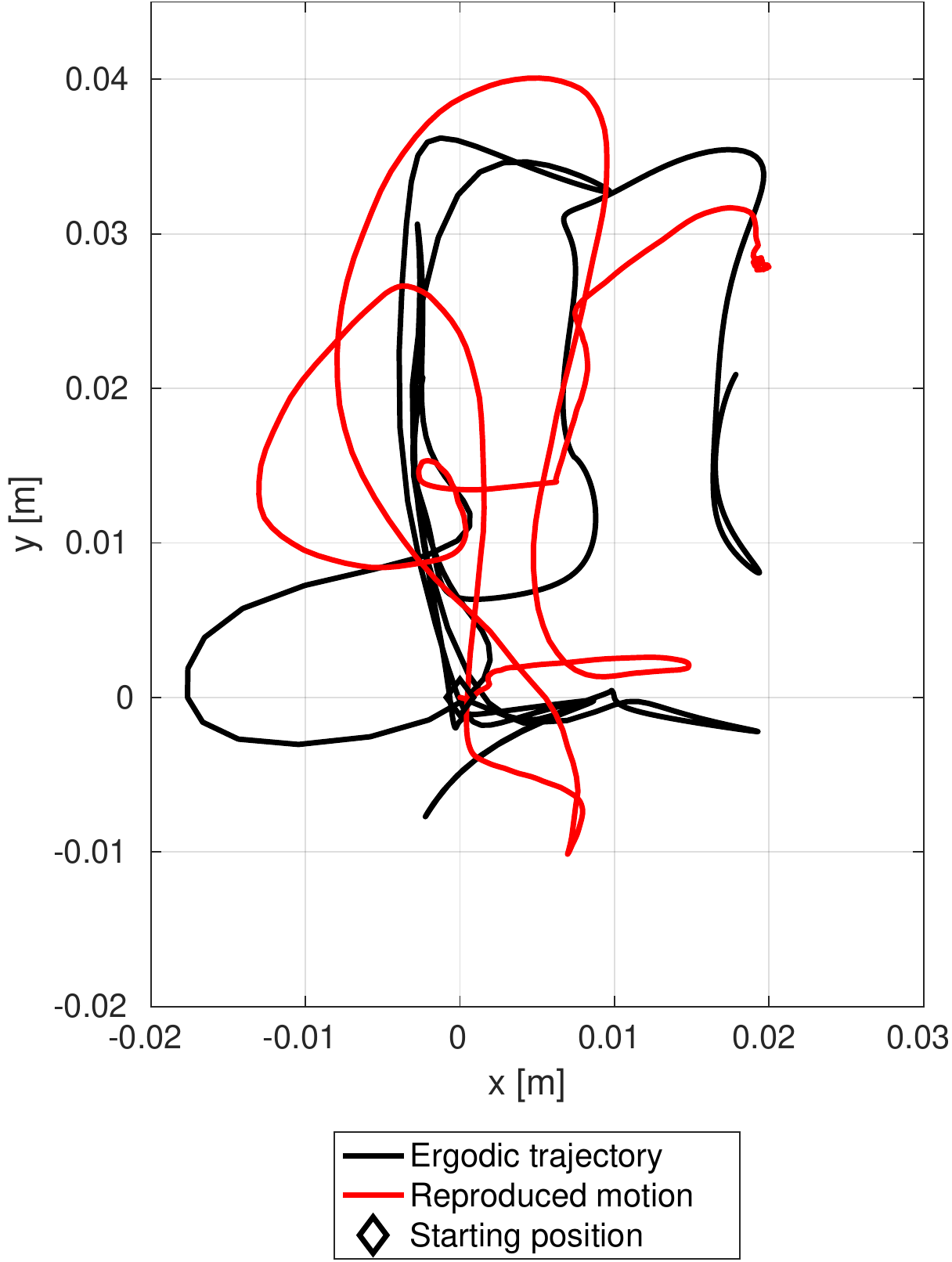}
   		\caption{}    		\label{fig:PiH_ERG_comp}
    \end{subfigure}
   	\caption{\small{Trajectory following performance of the controller on trajectories created by \ac{tshix} (a) and ergodic control (b).} \label{fig:PiH_comp}}
   	%\vspace{-1.5em}
\end{figure}

However, as seen in Fig.~\ref{fig:PiH_comp} even with the superposed forces the tool is not following the desired 
trajectory accurately. These 
trajectories were completed with stiffness values that were experimentally 
found to provide a good balance between trajectory following and compliance 
that facilitates the final insertion of the peg-in-hole task. The reason for 
this divergence lies presumably in the unknown friction forces of the robot's 
joints. When performing the demonstrations, the teachers also noticed that the 
joint configuration had a significant effect on the ease of moving the robot. 
In addition, feeding a Cartesian wrench profile recorded from a demonstration 
to the robot in pure force control did not produce any meaningful results. We 
assume this is due to static friction in the joints that prevents initiation of 
small motions. The same reason is why, in Table 
\ref{tab:samplingSizeComp}, the success percentage of \ac{tshix} did not reach 
$100\%$ despite increased sample size: in all sample sizes between 300 and 500, there was one error where the peg slid past the hole even if the planned trajectory covered it, and two errors where the peg entered the hole but did not slide all the way in, most likely due to insufficient compliance.
We hypothesize that better compensation of internal friction would allow trajectory following even with low stiffness and thus better facilitate the final insertion part.

\subsection{Learning a 3D Search Task}
\label{sec:3Dsearch}
\begin{figure}
	\centering
	\begin{subfigure}[tbp]{0.46\columnwidth}
		\includegraphics[width = \columnwidth]{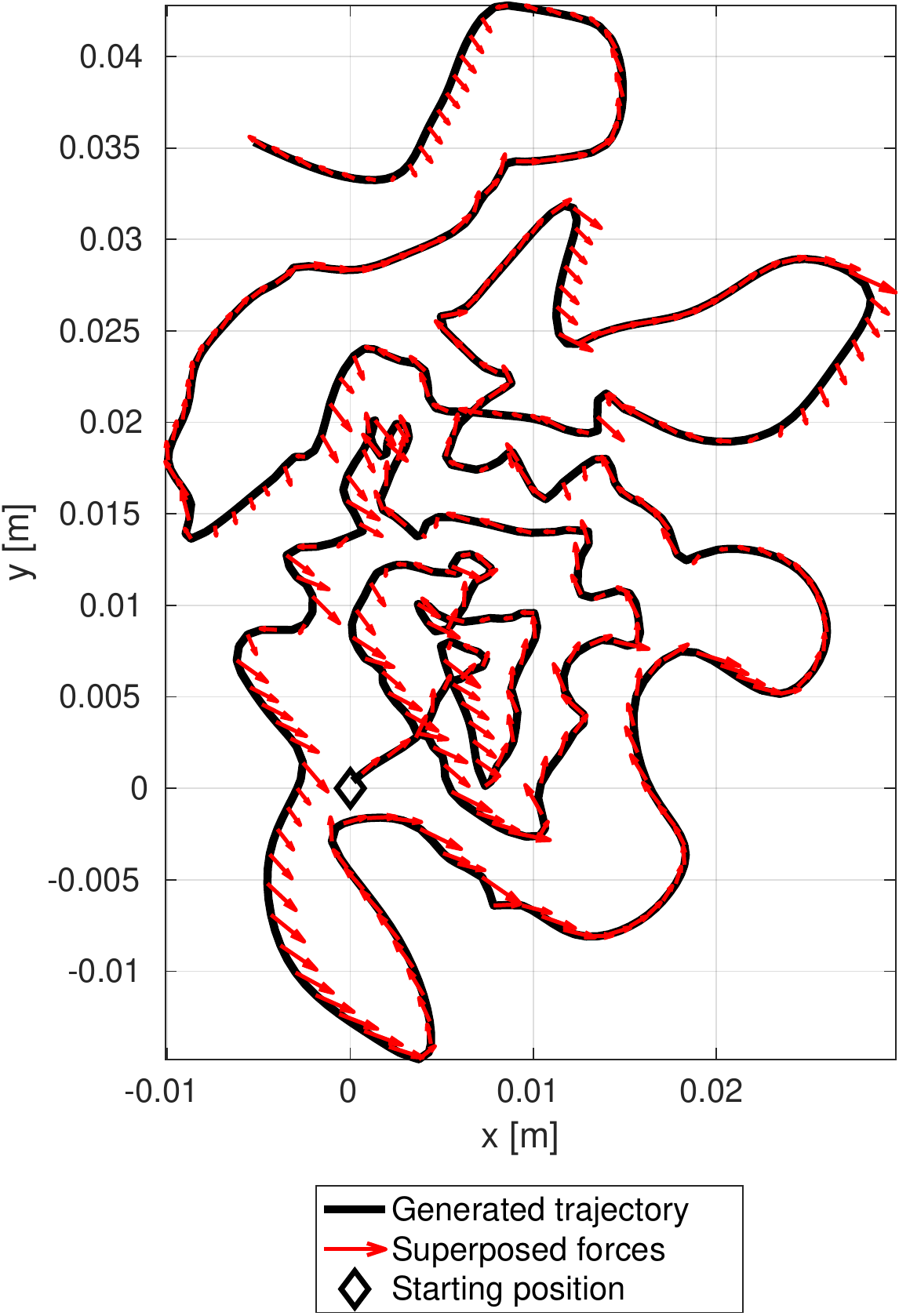}
		\caption{}
		\label{fig:PiH_predictedFrc}
		%\vspace{-1.5em}
	\end{subfigure}
\begin{subfigure}[tbp]{0.5\columnwidth}
        \includegraphics[width = \columnwidth]{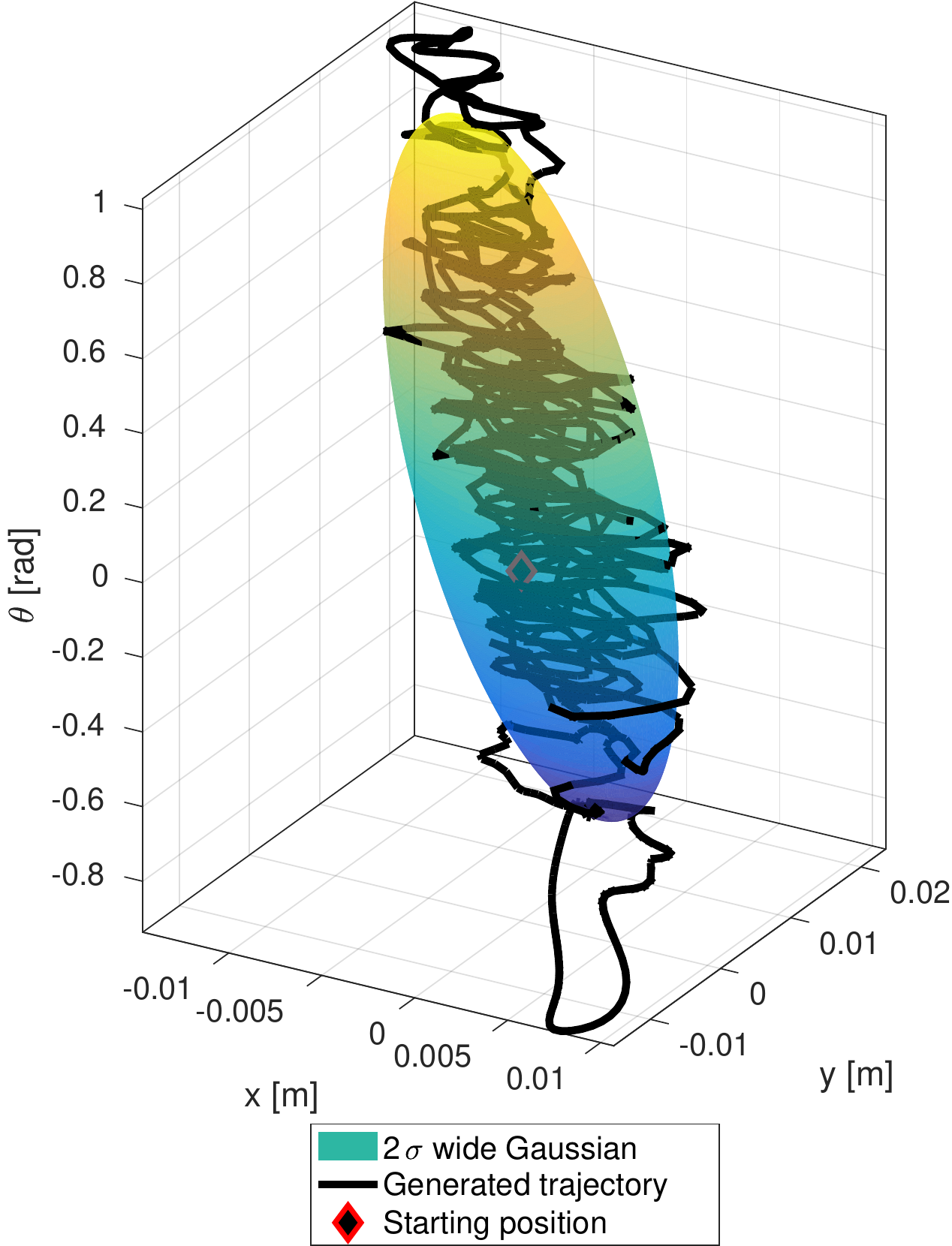}
   		\caption{}
    		\label{fig:Plug_trajectory}
    		%\vspace{-1em}
    	\end{subfigure}
    \caption{\small{(a) The superposed forces calculated from learned dynamics plotted along a generated search trajectory (b) Illustration of 3D trajectory created from 600 sample points.}}
    %\vspace{-1.5em}
\end{figure}

To investigate the scalability of the proposed method to higher dimensions, we conducted a three-dimensional search experiment. The task was to insert an electrical plug (Europlug) into a socket. For the 
socket, a 12V socket usually found in caravans and sheds was used because of its design: one hole is slightly larger, and 
there are valley shaped indentations around the holes, allowing an easier insertion when motion is compliant. To make the 
demonstrations and the learning process easier, the rotation around the x- and y-axes of the tool are restricted, so that 
the search was only performed in the positional space of the x-y-plane, as well as in rotation around the tool's z-axis. 
The demonstrations were conducted with closed eyes, and the starting pose of the plug was inside the socket close to a 
side wall. The orientation around the z-axis was chosen randomly, but still reasonable enough to allow the 
teacher to complete the task like an everyday situation. Because of the higher complexity of this 3D task, we compared 
learning the exploration distribution from either one or two demonstrations. The two demonstrations were performed starting from 
opposite ends of the upper half of the socket. From both a single demonstration and a 
combination of two demonstrations, the 3D exploration distribution (position in x and y, rotation around z) and the 6D task 
dynamics (position in x and y, rotation around z, forces in x and y, torques around z) were learned similarly to the 
peg-in-hole task, by fitting a single Gaussian distribution over the data. A trajectory was created by sampling 600 points, 
which was found to be a reasonable number for covering the 3D exploration distribution (see 
Fig.~\ref{fig:Plug_trajectory}). Subsequently the predicted interaction values from the environment (both forces and 
torques) for the given trajectory were calculated.

For the experiments, 15 starting positions were randomly chosen in the northern hemisphere of the socket, and for each 
starting position a different trajectory was generated from both one and two demonstrations. It was impossible to create an 
ergodic trajectory with Projection-based Trajectory Optimization in 3D, since the algorithm gets easily trapped in a local minimum which restricts the 
length of the created trajectory. In order to provide a baseline for comparison, random walk trajectories were created 
using the method proposed by Abu-Dakka et al.~\cite{abu2014solving}. The results are presented in 
Table~\ref{tab:result3Dtask}. We can observe that for the 3D task having two demonstrations significantly increases the 
success rate and even for the much harder socket task the method achieves good results.

\begin{table}[h]
    \centering
    \begin{tabular}{l c c c}
        \toprule
        Algorithm & Random Walk & 1-demo-\ac{tshix} & 2-demo-\ac{tshix} \\
        \midrule
        Success rate & 0/15 & 3/15 & 10/15 \\
        Success percentage & 0\% & 20\% & 67\% \\
        \bottomrule
    \end{tabular}
    \caption{\small{Success rates from plug-in-socket experiments from random walk and \ac{tshix} with either one or two demonstrations.}}
    \label{tab:result3Dtask}
\end{table}

\subsection{Discussion}
\label{sec:discussion}
The results show that the proposed method performs well if the demonstrations' starting positions cover the variability of the 
scenarios. However, some real world scenarios such as plug-and-socket for a European 230V AC socket are surprisingly 
difficult, even human teachers found it very hard to solve the task blindly. This indicates that some search problems may be infeasible in higher dimensions. On the other hand, if the 
demonstrated search is performed in a lower-dimensional subspace of a high-dimensional state space, the proposed method 
will automatically perform the search in the lower-dimensional space by virtue of employing the Gaussian linear model. 

We showed that while it is possible to use only one demonstration to learn a 
search if the task is sufficiently easy, for more complex tasks with higher 
dimensionality it is necessary to have more demonstrations in order to achieve 
a reasonable convergence region. Moreover, for multimodal or 
more complex tasks, the exploration distribution can be modeled as a \ac{gmm}. 
We originally used \ac{gmm}s, but initial experiments showed that only one Gaussian component was sufficient to model our experimental problems and more components lead to overfitting. Furthermore, a requirement for the 
demonstrations is that they need to be performed in worst case scenarios, so 
that they cover the variability of both exploration region and dynamics 
encountered in the task.

\section{CONCLUSIONS AND FUTURE WORK}
\label{sec:conclusions}

We presented a simple yet data efficient \ac{lfd} method teaching 
robots to imitate human search strategies for assembly tasks. The key idea is 
to first learn from human demonstrations both an exploration distribution and a 
dynamics 
model and then use these to generate candidate search trajectories covering 
said 
exploration 
distribution with superposed forces extracted from the dynamics model. We 
evaluated two different trajectory generating methods: \ac{tshix}, our own 
sampling based method, and ergodic control, a deterministic optimal path planner. 
The experimental results demonstrate that our method is data efficient as it 
learns 
to complete 2D and 3D search tasks from only one, respectively, two human 
demonstrations, a major improvement over current exception strategies that have 
been shown to only work in 2D. 

The method suffers from the same issue most \ac{lfd} methods do, mainly that it can only learn what it has seen. One remedy to this can be found in the research regarding Human-Robot Interaction, where a recent study investigated how to instruct human teachers to give informative demonstrations \cite{sena2018teaching}.

There are several interesting directions for future work. First of all, as we 
observed from demonstrations that humans tend to naturally 
perform 
their search in distinct phases and in no more than three dimensions, it stems 
to 
reason that it is more efficient to model high dimensional searches 
as a sequence of lower-dimensional ones with, \eg, the method presented in 
\cite{hagos2018seq}. Another interesting direction is to 
analyze whether the temporal correlations of the human search can be included 
in the exploration distribution. Furthermore, as experienced during the 
initial trials the 
quality of the human demonstrations varied substantially, raising an 
interesting question on how to 
develop teaching environment for both human and robot in order to get high 
quality demonstrations.

\addtolength{\textheight}{10cm}

\bibliographystyle{ieeetr}
\bibliography{biblio}
\addtolength{\textheight}{-10cm}

\end{document}